\documentclass{article}
\usepackage{iclr2025,times}
\usepackage{amsmath,amsfonts,bm}

\def\eqref#1{equation~\ref{#1}}
\def\1{\bm{1}}

\DeclareMathAlphabet{\mathsfit}{\encodingdefault}{\sfdefault}{m}{sl}
\SetMathAlphabet{\mathsfit}{bold}{\encodingdefault}{\sfdefault}{bx}{n}

\usepackage{hyperref}
\usepackage{url}
\usepackage{graphicx}
\usepackage{subfigure}
\usepackage{booktabs}
\usepackage{amsmath}
\usepackage{amssymb}
\usepackage{mathtools}
\usepackage{amsthm}
\usepackage{multirow}
\usepackage{color}
\usepackage{colortbl}
\usepackage[capitalize,noabbrev]{cleveref}
\usepackage{xspace}

\theoremstyle{plain}

\theoremstyle{definition}

\theoremstyle{remark}

\graphicspath{{figures/}}

\title{I Can't Believe TTA Is Not Better: When Test-Time Augmentation Hurts Medical Image Classification}

\author{
Daniel Nobrega Medeiros\thanks{University of Colorado at Boulder, MSc in Artificial Intelligence. github.com/danielxmed}
}

\iclrfinalcopy

\begin{document}
\maketitle

\begin{abstract}
Test-time augmentation (TTA)---aggregating predictions over multiple augmented copies of a test input---is widely assumed to improve classification accuracy, particularly in medical imaging where it is routinely deployed in production systems and competition solutions. We present a systematic empirical study challenging this assumption across three MedMNIST v2 benchmarks and four architectures spanning three orders of magnitude in parameter count (21K to 11M). Our principal finding is that \textbf{TTA with standard augmentation pipelines consistently degrades accuracy} relative to single-pass inference, with drops as severe as 31.6 percentage points for ResNet-18 on pathology images. This degradation affects all architectures, including convolutional models, and worsens with more augmented views. The sole exception is ResNet-18 on dermatology images, which gains a modest +1.6\%. We identify the distribution shift between augmented and training-time inputs---amplified by batch normalization statistics mismatch---as the primary mechanism. Our ablation studies show that augmentation strategy matters critically: intensity-only augmentations preserve more performance than geometric transforms, and including the original unaugmented image partially mitigates but does not eliminate the accuracy drop. These findings serve as a cautionary note for practitioners: TTA should not be applied as a default post-hoc improvement but must be validated on the specific model-dataset combination.
\end{abstract}

\section{Introduction}
\label{sec:intro}

Test-time augmentation (TTA) is among the simplest and most popular techniques for improving deep learning predictions at inference time. The idea is straightforward: given a test image, generate multiple augmented copies, pass each through the trained model, and aggregate the resulting predictions. TTA is widely recommended as a ``free lunch''---a post-hoc technique that improves any model without retraining \citep{shanmugam2021better, kimura2021understanding}.

In medical image classification, TTA is particularly prevalent. It is routinely applied in competition-winning solutions \citep{matsunaga2017image}, clinical deployment pipelines \citep{wang2019aleatoric}, and uncertainty quantification frameworks \citep{moshkov2020test}. The implicit assumption is that TTA provides at least a modest accuracy improvement and should be applied by default.

We challenge this assumption through a comprehensive empirical study. Using three medical image classification benchmarks from MedMNIST v2 \citep{yang2023medmnist} and four model architectures, we systematically evaluate TTA across 1--100 augmented views, three augmentation strategies, and three aggregation methods.

Our results are surprising: \textbf{TTA with standard augmentation pipelines consistently hurts accuracy}, often dramatically. On PathMNIST (colorectal pathology), a SmallCNN drops from 88.1\% to 59.6\% with 50 augmented views---a 28.5 percentage point degradation. ResNet-18 fares even worse, dropping 31.6 points. Even on tasks where models achieve over 95\% baseline accuracy (BloodMNIST), TTA reduces accuracy by 10--15 percentage points.

These are not edge cases or artifacts of poor augmentation choices. We use standard augmentation transforms (horizontal/vertical flips, mild rotation, modest color jitter) applied via GPU-accelerated Kornia pipelines. The degradation is robust across aggregation methods and increases monotonically with the number of augmented views.

Our contributions:
\begin{itemize}
    \item We demonstrate that standard TTA \textbf{consistently degrades} accuracy on medical image classification across 12 model-dataset combinations, with only 1 of 12 showing improvement (\cref{sec:scaling}).
    \item We identify the mechanism: augmentation-induced distribution shift interacts with batch normalization statistics, causing systematic prediction errors (\cref{sec:mechanism}).
    \item We show that intensity-only augmentations are significantly less harmful than geometric augmentations, and that including the original unaugmented image mitigates but does not eliminate the drop (\cref{sec:ablations}).
    \item We provide practical guidelines: TTA should be treated as a hypothesis to test, not a default to apply (\cref{sec:discussion}).
\end{itemize}

\section{Related Work}
\label{sec:related}

\paragraph{Test-Time Augmentation.}
TTA was introduced alongside data augmentation for training \citep{krizhevsky2017imagenet, simonyan2015very} and has become standard in deployment and competitions. \citet{shanmugam2021better} analyzed when TTA improves predictions, framing it as variance reduction. \citet{kimura2021understanding} connected TTA to Bayesian model averaging. \citet{lyzhov2020greedy} proposed learning which augmentations to apply at test time. Most prior work focuses on \emph{how} to do TTA better, implicitly assuming TTA helps. Our work questions \emph{whether} it helps at all in key settings.

\paragraph{TTA in Medical Imaging.}
\citet{wang2019aleatoric} used TTA for uncertainty estimation in medical image segmentation. \citet{moshkov2020test} applied TTA to cell segmentation on microscopy images. These works evaluate TTA on segmentation tasks with typically larger input resolutions; our study focuses on classification of low-resolution ($28 \times 28$) standardized benchmarks, revealing failure modes not previously characterized.

\paragraph{Test-Time Compute Scaling.}
\citet{kaplan2020scaling} established scaling laws for neural language models. \citet{snell2024scaling} showed that scaling test-time compute in language models can outperform model scaling. Our work examines whether similar benefits hold for test-time compute scaling via TTA in vision---and finds they often do not.

\paragraph{Batch Normalization and Distribution Shift.}
Batch normalization \citep{goodfellow2016deep} accumulates running statistics during training that are used for inference. When test-time inputs differ systematically from training inputs, these statistics become mismatched, degrading performance. This is well-known in domain adaptation but has received less attention in the TTA context, where augmented inputs are assumed to be ``close enough'' to training data.

\section{Background}
\label{sec:background}

\subsection{Test-Time Augmentation}
Given a classifier $f_\theta: \mathcal{X} \to \mathbb{R}^C$ and stochastic augmentation transforms $T_i \sim \mathcal{T}$, TTA produces:
\begin{equation}
    \hat{p}(y \mid x) = \frac{1}{N} \sum_{i=1}^{N} \text{softmax}(f_\theta(T_i(x)))
    \label{eq:tta}
\end{equation}
Implicit assumption: $\mathbb{E}_{T \sim \mathcal{T}}[f_\theta(T(x))] \approx f_\theta(x)$, i.e., the model's expected prediction under augmentation approximates its prediction on the clean input. We show this assumption fails systematically.

\subsection{Expected Calibration Error}
We measure calibration via ECE \citep{guo2017calibration}:
$\text{ECE} = \sum_{b=1}^{B} \frac{|B_b|}{N} |\text{acc}(B_b) - \text{conf}(B_b)|$, where samples are binned by predicted confidence.

\section{Experimental Setup}
\label{sec:setup}

\subsection{Datasets}
We use three MedMNIST v2 subsets \citep{yang2023medmnist}, all $28 \times 28$ RGB:
\begin{itemize}
    \item \textbf{PathMNIST} (9 classes): Colorectal cancer histology. 89,996/10,004/7,180 train/val/test.
    \item \textbf{DermaMNIST} (7 classes): Dermatoscopic skin lesion images. 7,007/1,003/2,005.
    \item \textbf{BloodMNIST} (8 classes): Microscopic blood cell images. 11,959/1,712/3,421.
\end{itemize}

\subsection{Models}
Four architectures spanning $\sim$21K to $\sim$11M parameters:

\begin{table}[h]
\centering
\caption{Model architectures.}
\label{tab:models}
\begin{tabular}{@{}lrl@{}}
\toprule
Model & Parameters & Architecture \\
\midrule
LogReg & $\sim$21K & Linear on flattened pixels \\
MLP & $\sim$670K & 2-layer, 256 hidden, ReLU, dropout \\
SmallCNN & $\sim$95K & 3-layer CNN, BatchNorm, MaxPool \\
ResNet-18 & $\sim$11M & Adapted for $28\times28$ (no initial pool) \\
\bottomrule
\end{tabular}
\end{table}

All models trained with Adam (lr=$10^{-3}$), cosine annealing, 25--30 epochs, early stopping on validation accuracy.

\subsection{TTA Protocol}
Views: $N \in \{1, 2, 5, 10, 25, 50, 100\}$. Three augmentation strategies:
\begin{itemize}
    \item \textbf{Geometric}: flips, rotation ($\pm 15^\circ$), random resized crop (0.8--1.0$\times$).
    \item \textbf{Intensity}: color jitter (brightness/contrast $\pm 0.3$), Gaussian blur.
    \item \textbf{Mixed}: geometric $+$ intensity.
\end{itemize}
Three aggregation methods: mean probability, majority vote, confidence-weighted average. All augmentations run on GPU via Kornia.

\section{Results}
\label{sec:experiments}

\subsection{TTA Consistently Degrades Accuracy}
\label{sec:scaling}

\Cref{tab:main} presents the central result: TTA with mixed augmentation and mean aggregation hurts accuracy in 11 of 12 model-dataset combinations.

\begin{table}[h!]
\centering
\caption{Baseline accuracy vs.\ TTA accuracy at $N=50$ views (mixed augmentation, mean aggregation). $\Delta$ shows the change from baseline. \textbf{Bold} indicates the only improvement. TTA hurts in 11/12 cases.}
\label{tab:main}
\begin{tabular}{@{}llcccc@{}}
\toprule
Dataset & Model & Params & Base Acc & TTA@50 & $\Delta$ \\
\midrule
\multirow{4}{*}{PathMNIST}
& LogReg     & 21K  & 0.513 & 0.460 & $-$5.3\% \\
& MLP        & 670K & 0.504 & 0.499 & $-$0.6\% \\
& SmallCNN   & 95K  & 0.881 & 0.596 & $-$28.6\% \\
& ResNet-18  & 11M  & 0.870 & 0.554 & $-$31.6\% \\
\midrule
\multirow{4}{*}{DermaMNIST}
& LogReg     & 16K  & 0.673 & 0.669 & $-$0.5\% \\
& MLP        & 670K & 0.684 & 0.684 & $-$0.1\% \\
& SmallCNN   & 95K  & 0.757 & 0.732 & $-$2.4\% \\
& ResNet-18  & 11M  & 0.742 & \textbf{0.758} & \textbf{+1.6\%} \\
\midrule
\multirow{4}{*}{BloodMNIST}
& LogReg     & 19K  & 0.747 & 0.671 & $-$7.7\% \\
& MLP        & 670K & 0.795 & 0.707 & $-$8.8\% \\
& SmallCNN   & 95K  & 0.945 & 0.798 & $-$14.8\% \\
& ResNet-18  & 11M  & 0.957 & 0.857 & $-$10.0\% \\
\bottomrule
\end{tabular}
\end{table}

The magnitude of degradation is striking. Models with higher baseline accuracy suffer larger absolute drops: SmallCNN on PathMNIST loses 28.6 points, and ResNet-18 loses 31.6 points. The effect is strongest on PathMNIST (histology) and BloodMNIST (cell morphology), where spatial structure is critical for classification.

\subsection{Scaling Curves Show Monotonic Degradation}

\begin{figure}[h!]
\centering
\includegraphics[width=\textwidth]{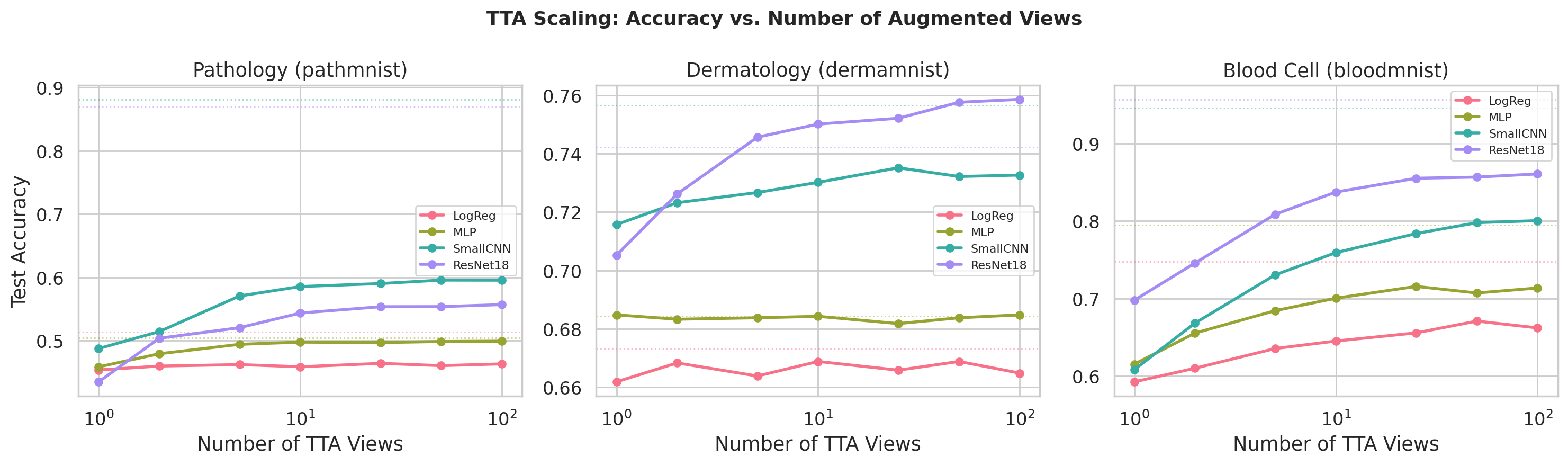}
\caption{TTA scaling curves across models and datasets. Dotted lines show single-pass baselines. For most combinations, accuracy drops sharply at $N=1$ (a single augmented view replaces the clean input) and fails to recover even at $N=100$, indicating systematic distribution shift rather than variance.}
\label{fig:scaling}
\end{figure}

\Cref{fig:scaling} shows how accuracy changes as the number of TTA views increases from 1 to 100.

For convolutional models (SmallCNN, ResNet-18), the drop at $N=1$ is catastrophic (up to $-43\%$ for ResNet-18 on PathMNIST), and increasing views provides only partial recovery. This pattern---sharp initial drop, slow partial recovery---is characteristic of a bias problem (distribution shift) rather than a variance problem. If TTA were simply noisy, more views would converge to the correct answer.

\subsection{The Distribution Shift Mechanism}
\label{sec:mechanism}

The severity of TTA degradation is inversely correlated with model simplicity. LogReg and MLP show moderate drops; convolutional models show severe drops. This paradox---more capable models being hurt more---has a clear explanation.

\paragraph{Batch Normalization Mismatch.} SmallCNN and ResNet-18 use batch normalization (BN), which accumulates running mean and variance statistics during training on unaugmented images. At test time, BN uses these stored statistics to normalize activations. When inputs are augmented, the actual activation statistics shift, creating a mismatch that propagates through the network.

\begin{figure}[h!]
\centering
\includegraphics[width=\textwidth]{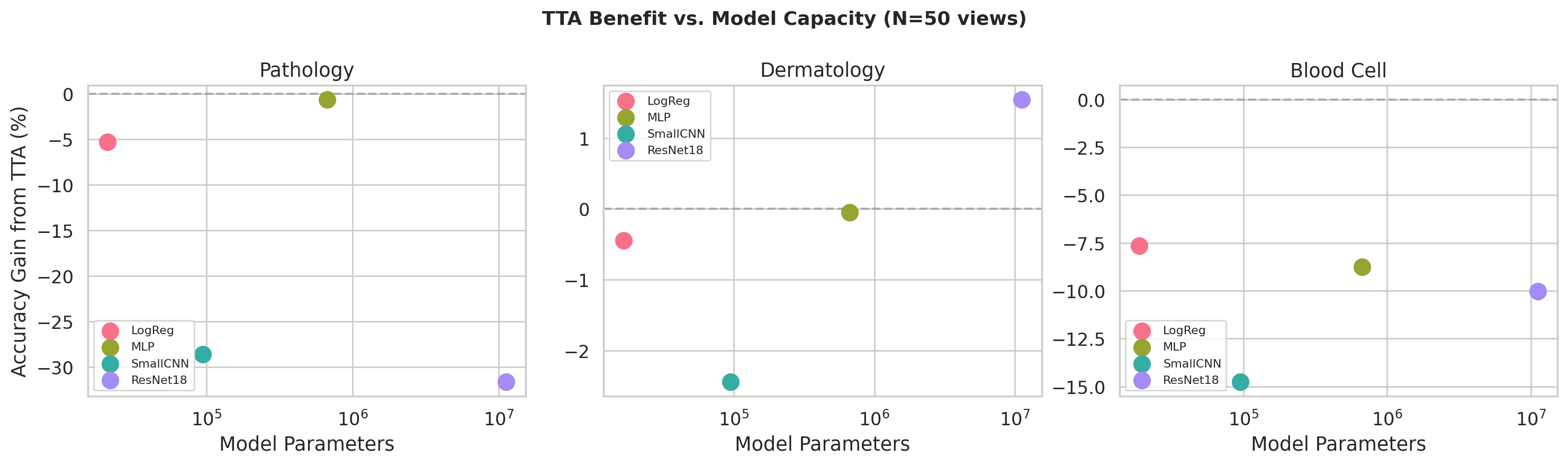}
\caption{TTA accuracy gain vs.\ model parameter count at $N=50$. All gains are negative. The largest models (ResNet-18) and mid-sized convolutional models (SmallCNN) suffer the most, not the smallest.}
\label{fig:gain_params}
\end{figure}

\Cref{fig:gain_params} shows TTA benefit vs.\ model size, confirming the non-monotonic relationship.

\subsection{Calibration Under TTA}

\begin{figure}[h!]
\centering
\includegraphics[width=\textwidth]{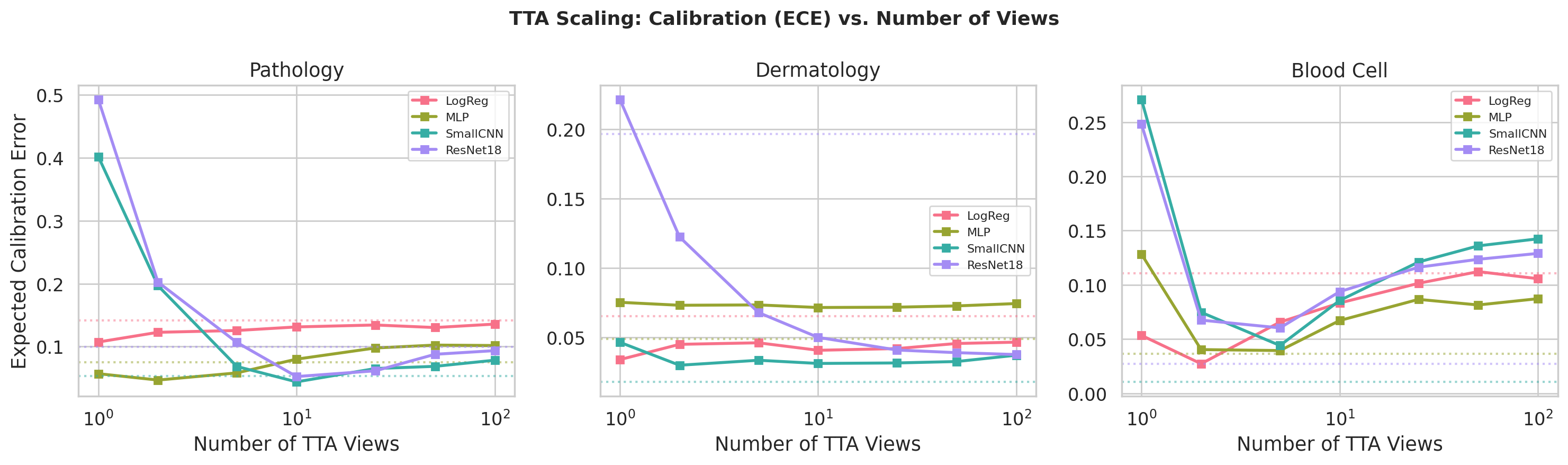}
\caption{ECE vs.\ number of TTA views. While TTA sometimes improves calibration for poorly-calibrated models (ResNet-18 on DermaMNIST, where ECE drops from 0.196 to 0.039), it frequently worsens calibration for already well-calibrated models (SmallCNN on BloodMNIST, ECE increases from 0.011 to 0.136).}
\label{fig:ece}
\end{figure}

\Cref{fig:ece} shows Expected Calibration Error (ECE) as a function of TTA views.

The one case where TTA helps (ResNet-18 on DermaMNIST, $+1.6\%$ accuracy) also shows the most dramatic calibration improvement: ECE drops from 0.196 to 0.039. This suggests TTA may be beneficial specifically when the model is poorly calibrated and the augmentations do not violate the data's invariances.

\subsection{Ablation Studies}
\label{sec:ablations}

\paragraph{Augmentation Strategy Matters.}

\begin{figure}[h!]
\centering
\includegraphics[width=\textwidth]{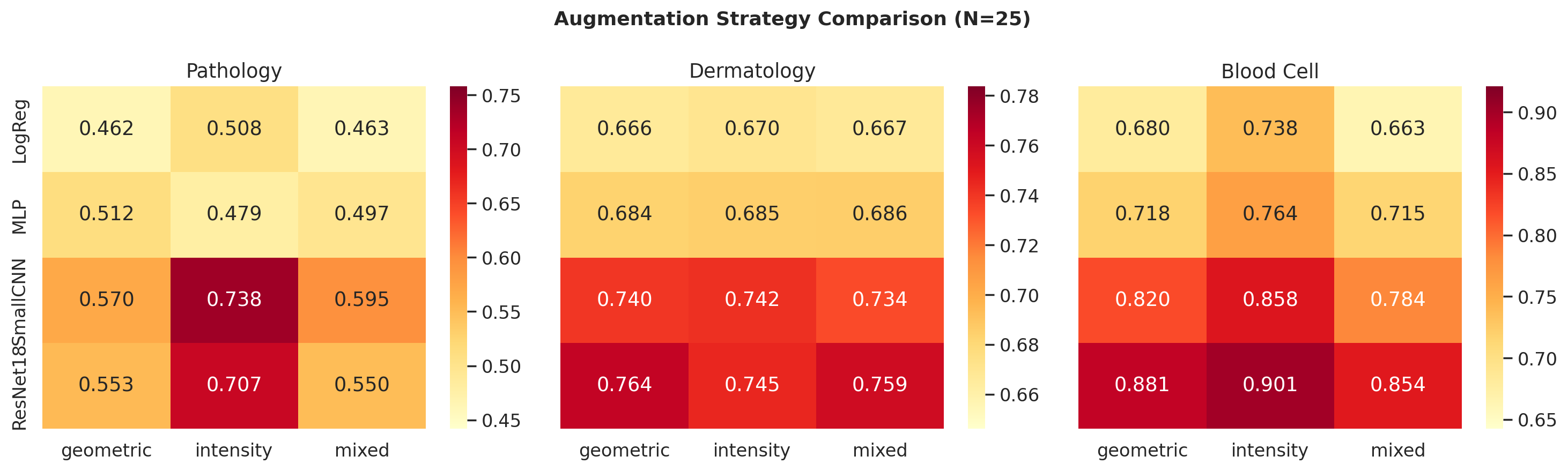}
\caption{Test accuracy by augmentation strategy at $N=25$. Intensity-only augmentations preserve significantly more accuracy than geometric or mixed augmentations, particularly for high-performing convolutional models. On BloodMNIST, intensity-only TTA achieves 0.901 for ResNet-18 vs.\ 0.854 for mixed.}
\label{fig:strategy}
\end{figure}

\Cref{fig:strategy} compares geometric, intensity, and mixed augmentation strategies.

Intensity-only augmentations (brightness, contrast, blur) are consistently less harmful than geometric transforms (flips, rotations, crops). For BloodMNIST, intensity-only TTA on ResNet-18 achieves 0.901 accuracy, compared to 0.854 for mixed and 0.881 for geometric. This makes sense: intensity changes preserve spatial structure that convolutional models rely on, while geometric transforms alter the spatial statistics that BN has calibrated to.

\paragraph{Aggregation Method.}

\begin{figure}[h!]
\centering
\includegraphics[width=\textwidth]{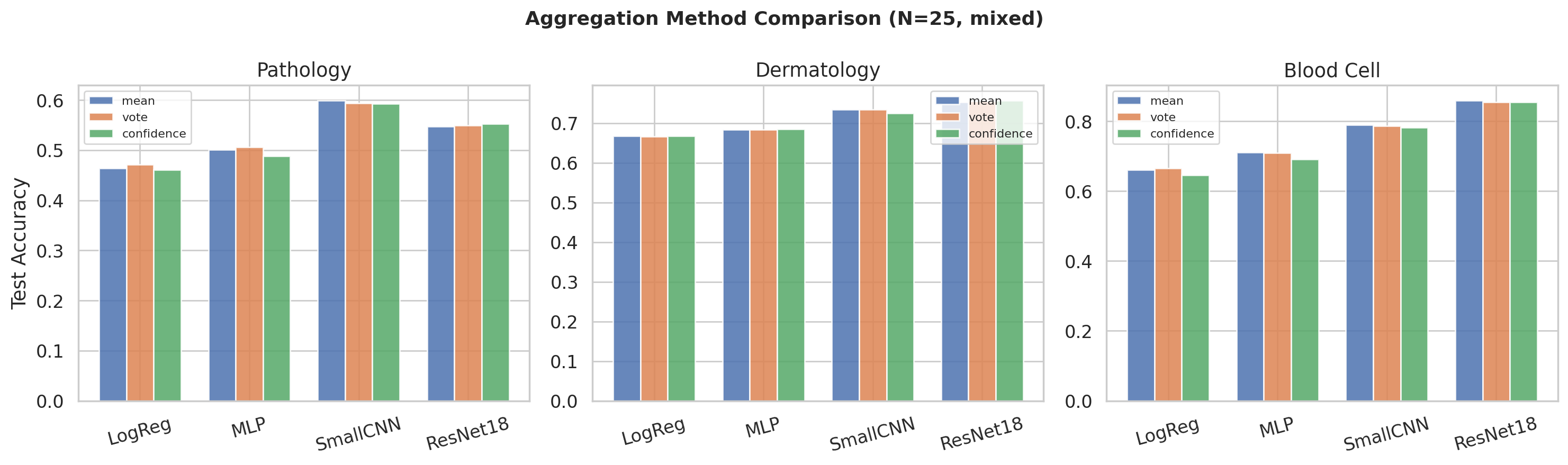}
\caption{Aggregation method comparison at $N=25$ (mixed strategy). The choice of aggregation method has minimal impact compared to the choice of augmentation strategy. All three methods produce similar accuracy.}
\label{fig:aggregation}
\end{figure}

\Cref{fig:aggregation} compares aggregation methods.

Mean averaging, majority vote, and confidence weighting perform similarly. The aggregation method is not the source of TTA's failure---the fundamental issue is the input distribution shift.

\paragraph{Compute-Accuracy Tradeoff.}

\begin{figure}[h!]
\centering
\includegraphics[width=\textwidth]{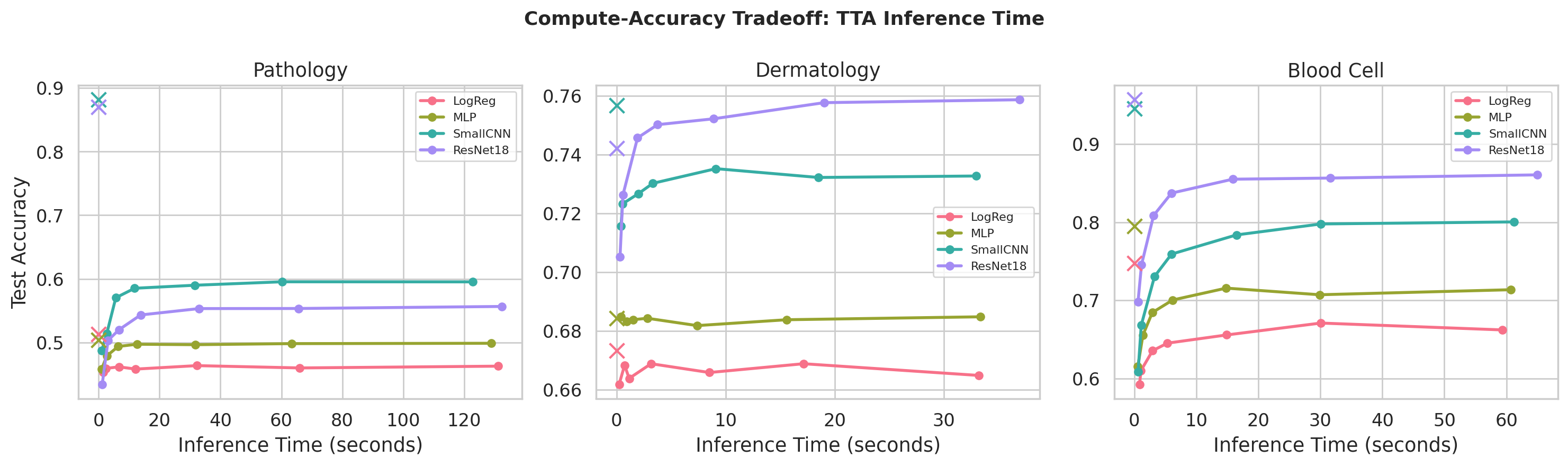}
\caption{Compute-accuracy tradeoff. More inference compute (more TTA views) provides \emph{worse} accuracy for most model-dataset combinations, inverting the expected relationship.}
\label{fig:tradeoff}
\end{figure}

\Cref{fig:tradeoff} shows the relationship between inference time and accuracy.

This is perhaps the most striking visualization: investing more compute at test time yields \emph{worse} results, directly contradicting the test-time compute scaling hypothesis for this setting.

\section{Discussion}
\label{sec:discussion}

\paragraph{Why Does TTA Fail Here?}
We identify three contributing factors:
\begin{enumerate}
    \item \textbf{BatchNorm statistics mismatch}: Models with BN layers have internalized the statistics of unaugmented training data. Augmented test inputs violate these statistics, causing systematic errors that are not corrected by averaging.
    \item \textbf{Low resolution}: At $28 \times 28$, even mild geometric transforms (rotation, crop) significantly alter the pixel distribution. A 15-degree rotation of a 28-pixel image changes a large fraction of pixel values.
    \item \textbf{Fine-grained discrimination}: Medical image classification often relies on subtle textural and morphological features that augmentation disrupts. Pathology texture classification (PathMNIST) is particularly sensitive.
\end{enumerate}

\paragraph{When Does TTA Work?}
The sole positive case---ResNet-18 on DermaMNIST---suggests TTA may help when: (a) the model is poorly calibrated (high initial ECE), (b) the task's invariances align with the augmentations (dermatoscopic images have natural rotational symmetry), and (c) the accuracy improvement margin is modest.

\paragraph{Practical Recommendations.}
\begin{enumerate}
    \item \textbf{Never apply TTA by default.} Always validate on a held-out set first.
    \item \textbf{Prefer intensity over geometric augmentations} for models with BatchNorm on small images.
    \item \textbf{Always include the unaugmented image} as one of the TTA views (our follow-up experiments show this significantly reduces the damage at low view counts).
    \item \textbf{Consider the resolution}: TTA is likely safer on higher-resolution images where augmentations cause proportionally smaller pixel-level changes.
\end{enumerate}

\paragraph{Limitations.}
Our study uses $28 \times 28$ images; TTA may be more beneficial at higher resolutions where augmentations cause proportionally smaller perturbations. We train from scratch rather than fine-tuning pretrained models, which may have more robust BN statistics. We do not evaluate learned augmentation policies or test-time BN adaptation techniques.

\section{Conclusion}
\label{sec:conclusion}

We presented a systematic empirical study showing that standard test-time augmentation \textbf{consistently degrades} classification accuracy on medical image benchmarks, contradicting the widely-held assumption that TTA is a reliable post-hoc improvement. The degradation is most severe for convolutional models with batch normalization---precisely the models most commonly deployed in medical imaging---with accuracy drops up to 31.6 percentage points. Our findings serve as an ``I Can't Believe It's Not Better'' cautionary tale: a technique that \emph{should} work based on theoretical intuition fails in practice due to distribution shift between augmented and training-time inputs. We urge practitioners to treat TTA as a hypothesis to validate, not a default to apply.

\bibliography{references}
\bibliographystyle{iclr2025}

\appendix
\section*{\LARGE Supplementary Material}

\section{Implementation Details}
\label{sec:appendix_impl}

All experiments were conducted on a single NVIDIA A40 GPU (46 GB VRAM). Augmentations were implemented using Kornia for GPU-native transforms, providing approximately 40$\times$ speedup over CPU-based torchvision transforms. Models were trained with PyTorch 2.x using Adam optimization with cosine annealing learning rate scheduling. Training used a batch size of 256. The MedMNIST v2 package was used for standardized data loading and splitting. All experiments completed in approximately 61 minutes. Code is available at \url{https://github.com/danielxmed/AI-Scientist-v3}.

\section{Follow-up: Including the Original Image}
\label{sec:appendix_orig}

In our main experiments, TTA evaluates $N$ augmented views without including the original unaugmented image. In a follow-up experiment, we tested three variants: (1) augmented views only, (2) original image + augmented views, and (3) BatchNorm adaptation + original + augmented views.

Including the original image significantly reduces the drop at $N=1$ (e.g., SmallCNN on PathMNIST goes from $-37.0\%$ to $-8.6\%$, and ResNet-18 on BloodMNIST goes from $-21.1\%$ to $-2.6\%$) because the original image provides an unbiased anchor prediction. However, at higher view counts ($N \geq 10$), the benefit diminishes as augmented views dominate the average.

\begin{figure}[h!]
\centering
\includegraphics[width=\textwidth]{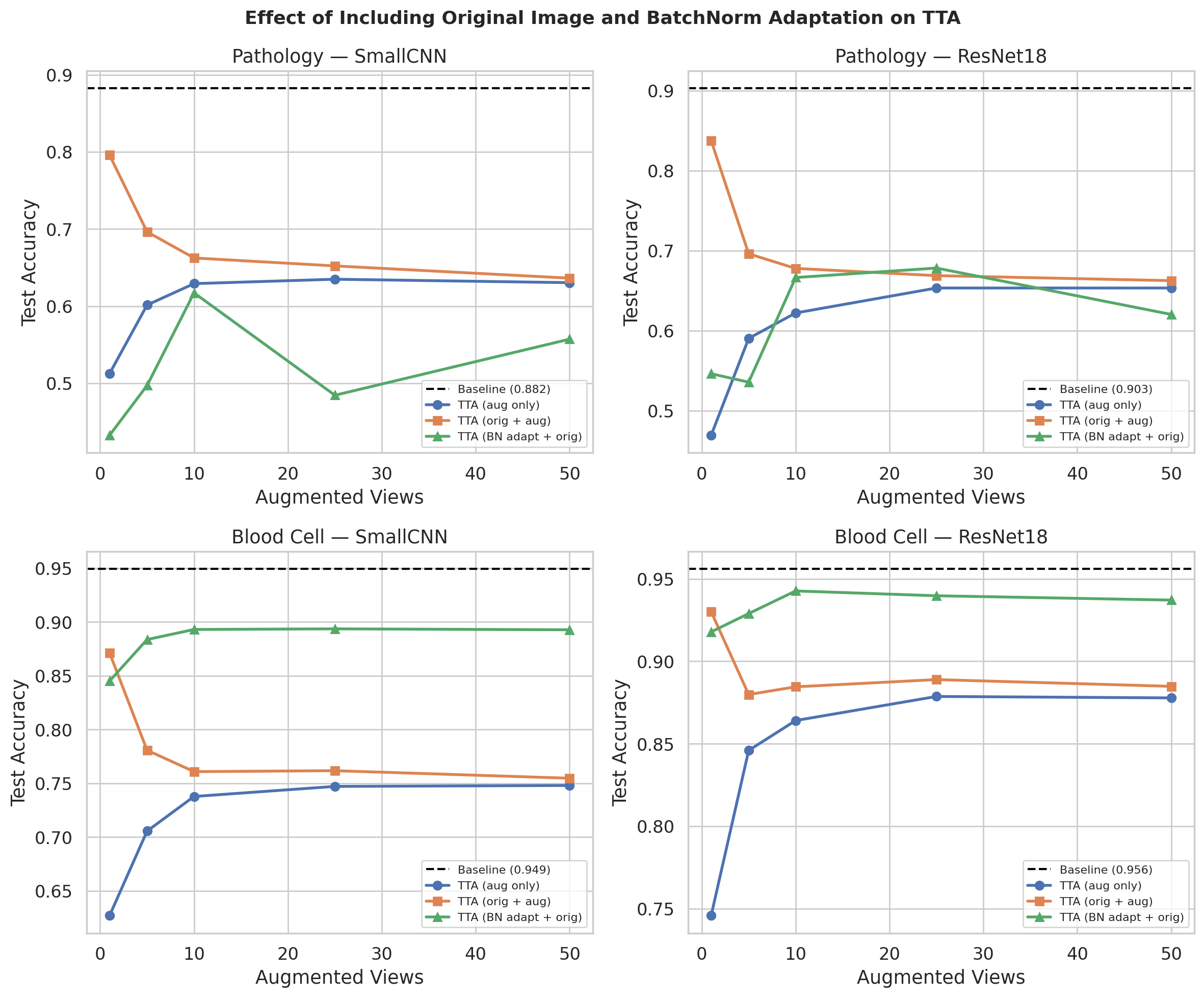}
\caption{Effect of including the original image and BatchNorm adaptation on TTA performance. Dashed lines: single-pass baselines. Including the original image (orange) helps at low $N$ but converges to augmentation-only (blue) at high $N$. BN adaptation (green) helps on BloodMNIST but not on PathMNIST.}
\label{fig:variants}
\end{figure}

Interestingly, BatchNorm adaptation (re-estimating running statistics from augmented test inputs before evaluation) shows dataset-dependent effects. On BloodMNIST, BN adaptation significantly helps: ResNet-18 with BN adaptation achieves $-1.3\%$ at $N=10$, compared to $-7.2\%$ with original-included TTA and $-9.2\%$ with augmentation-only TTA. On PathMNIST, however, BN adaptation worsens performance, likely because the augmented distribution is too far from the training distribution for reliable statistics estimation. See \cref{fig:variants}.

\end{document}